\documentclass[submission,copyright]{eptcs}
\usepackage{breakurl}
\def\titlerunning{The optimality of coarse categories in decision-making and
information storage}
\def\authorrunning{M. Mandler}

\begin{document}

\title{The optimality of coarse categories in decision-making and
information storage}
\author{Michael Mandler 
\institute{Department of Economics \and 
Royal Holloway College, University of London \and 
Egham, United Kingdom}}
\maketitle

\begin{abstract}
An agent who lacks preferences and instead makes decisions using criteria
that are costly to create should select efficient sets of criteria, where
the cost of making a given number of choice distinctions is minimized. \
Under mild conditions, efficiency requires that binary criteria with only
two categories per criterion are chosen. \ When applied to the problem of
determining the optimal number of digits in an information storage device,
this result implies that binary digits (bits) are the efficient solution,
even when the marginal cost of using additional digits declines rapidly to
0. \ This short paper pays particular attention to the symmetry conditions
entailed when sets of criteria are efficient.
\end{abstract}

\section{Introduction}

Suppose that agents, rather than forming a separate preference judgment for
each pair of alternatives, make decisions using \textit{criteria}. \ A
criterion orders a small number of categories, each of which consists of
many alternatives. \ The potential of a criterion to order alternatives
within another criterion's categories allow sets of criteria to generate
large numbers of choice distinctions. \ If an agent has objective
preferences that can be inferred from a large set of sufficiently
discriminating criteria, the agent will be better off if more of the
criterion orderings are discovered: the agent will then be able to determine
the optimal allocation from more choice sets. \ The uncovering of more
criterion discriminations is costly, however, and we therefore consider 
\textit{efficient} points where the cost of making a given number of choice
distinctions is minimized.

This optimization problem seems to lead to a trade-off. \ Given a number of
choice distinctions, an agent could either use a large set of coarse
criteria (criteria with only a small number of categories) or a small set of
finer, more discriminating criteria. \ We show under mild conditions that
large sets of coarse criteria always lead to reductions in decision-making
costs. \ Binary criteria with only two categories per criterion therefore
provide the only efficient arrangement. \ Under mild restrictions on how
criteria are aggregated into decisions, binary criteria lead to \textit{%
rational} choice functions, where decisions are determined by a complete and
transitive binary relation.

We apply our model to the problem of determining the optimal number of
digits in an information storage device. \ We show that, even if the
marginal cost of additional digits declines rapidly to 0, binary digits
(bits) offer the efficient solution.

In this short paper, we pay particular attention to the symmetry conditions
that are entailed when sets of criteria are efficient. \ A full working
paper [2] is available on-line.

\section{An outline of the model}

A \textit{criterion} $C_{i}$ is an asymmetric binary relation on a domain of
alternatives $X$ and a \textit{set of criteria} is denoted $\mathcal{C}%
=\{C_{1},...,C_{N}\}$. \ Two alternatives $x$ and $y$ are deemed $C_{i}$%
-equivalent if $x$ and $y$ share the same set of $C_{i}$-superior
alternatives and the same set of $C_{i}$-inferior alternatives (see [1]). \
A $C_{i}$-\textit{category} is a maximal set of $C_{i}$-equivalent
alternatives and $e(C_{i})$ denotes the number of $C_{i}$-categories. \ The 
\textit{discrimination vector} of $\mathcal{C}$ is $(e(C_{1}),...,e(C_{N}))$%
. \ A $C_{i}$ is \textit{coarser} than $C_{i}^{\prime }$ if $%
e(C_{i})<e(C_{i}^{\prime })$.

Let $c$ be a choice function on a domain of finite subsets of $X$. \ Two
alternatives $x$ and $y$ are in the same \textit{choice class} of $c$ if $c$
treats them interchangeably: first, when $x$ is chosen and $y$ is available
then $y$ is chosen too, and second, if $x$ but not $y$ is available then $x$
is chosen if and only if, when $y$ is available and not $x$, $y$ is chosen.

A choice function $c$ \textit{uses }$\mathcal{C}$, denoted $(\mathcal{C},c)$%
, if $c$ does not make distinctions that are not already present in the
criteria: for each set of alternatives $A$ that contains only alternatives
that are in the same $C_{i}$-category, $i=1,...,N$, there is a choice class
of $c$ that contains $A$.

Let $\kappa (C_{i})$ denote the \textit{cost of criterion} $C_{i}$. \ We
assume $\kappa (C_{i})$ is determined by the number of $C_{i}$-categories
and therefore also write $\kappa (e)$ to denote the cost of a $C_{i}$ with $%
e $ categories. \ We assume that the \textit{cost of a set of criteria}, $%
\kappa \lbrack \mathcal{C}]$, equals the sum of the costs of the criteria in 
$\mathcal{C}$. \ Letting $n(c)$ be the number of choice classes in $c$, a
pair $(\mathcal{C},c)$ is \textit{more efficient}\ than the pair $(\mathcal{C%
}^{\prime },c^{\prime })$ if $n(c)\geq n(c^{\prime })$ and $\kappa \lbrack 
\mathcal{C}]\leq \kappa \lbrack \mathcal{C}^{\prime }]$, with at least one
strict inequality, and $(\mathcal{C},c)$ is \textit{efficient} if there does
not exist a $(\mathcal{C}^{\prime },c^{\prime })$ that is more efficient
than $(\mathcal{C},c)$.

The fundamental advantage of criteria is that each criterion can
discriminate within the categories of other criteria. \ Given constraints
that specify that criterion $C_{i}$ can have no more than $e_{i}$ categories
(and assuming that $|X|$ is sufficiently large), we can find a $(\mathcal{C}%
,c)$ such that (i) there is a partition of $X$ with $\prod_{i=1}^{N}e_{i}$
cells such that $x$ and $y$ are in distinct cells if and only if they lie in
different $C_{i}\,$-categories for at least one $i$ and (ii) each cell of
this partition forms a choice class of $c$. \ Subject to the $e_{i}$\
constraints, this $(\mathcal{C},c)$ maximizes $n(c)$ and accordingly we
define $(\mathcal{C},c)$ to \textit{maximally discriminate} if $n(c)=\min %
\left[ \prod_{i=1}^{N}e(C_{i}),\left\vert X\right\vert \right] $.

\section{Main results}

(1) \ Since criteria with only one category make no discriminations and
require no decisions, we assume they are costless. \ To compare a $(\mathcal{%
C},c)$ and $(\mathcal{C}^{\prime },c^{\prime })$ that have the same number
of costly categories, suppose that $\sum_{i=1}^{N}\left( e(C_{i})-1\right)
=\sum_{i=1}^{N^{\prime }}\left( e(C_{i}^{\prime })-1\right) $. \ Assume also
that either (i) the marginal cost of categories is increasing and the
smaller of $n(c)$ and $n(c^{\prime })$ is less than the cardinality of $X$
or (ii) the marginal costs of categories is strictly increasing. \ We show
that if $\mathcal{C}$ has greater proportions of coarser criteria than does $%
\mathcal{C}^{\prime }$ and if $(\mathcal{C},c)$ maximally discriminates,
then $(\mathcal{C},c)$ is more efficient than $(\mathcal{C}^{\prime
},c^{\prime })$.

(2) \ Fix a set of domains that, for each finite $m$, contains a $X$ with $m$
elements and call a domain \textit{admissible} if it is drawn from this set.
\ Then, every efficient $(\mathcal{C},c)$ where the domain is admissible has
a $\mathcal{C}$ that contains only binary criteria if and only if $\kappa
(e)>\kappa (2)\left\lceil \log _{2}e\right\rceil $ for all integers $e>2$.

Thus the cost of $e$ categories can rise as slowly as $\log _{2}e$ -- in
which case the marginal cost of categories descends to $0$ -- and still the
only efficient arrangement is for all criteria to be binary.

(3) \ We apply the result in (2) to information storage. \ Suppose we wish
to store some integer between $1$ and $n$ using $N$ $k$-ary digits and that
the cost of storage equals $\kappa (k)N$. \ We show that, for all positive
integers $n$, binary digits are the minimum-cost storage method if and only
if $\kappa (k)>\kappa (2)\left\lceil \log _{2}k\right\rceil $ for all
integers $k>2$.

(4) \ We specify axioms for how to aggregate sets of criteria into choice
functions that generalize weighted voting. \ Suppose that the choice
function $c$ in the pair $(\mathcal{C},c)$ satisfies these axioms, that $%
\mathcal{C}$ contains only binary criteria, and that $c$ satisfies the
following Condorcet rule: if there is a $x$ in a choice set $A$ that is
chosen by $c$ from all $\{x,y\}$ with $y\in A$ then $x$ is chosen from $A$
too. \ Then $c$ makes selections that maximize a complete and transitive
binary relation. \ Given (2), we conclude that in a broad range of cases,
efficient decision-making is rational.

\section{Symmetry and maximal categorization}

Maximal discrimination is necessary for decision-making efficiency since
otherwise $n(c)$ could be increased without an increase in costs. \ The key
feature required for a $(\mathcal{C},c)$ to maximally discriminate is that
the following property of $\mathcal{C}$, called \textit{maximal
categorization}, is satisfied: the \textit{discrimination partition} $%
\mathcal{P}$\ of $X$ that places $x$ and $y$ in distinct cells if and only
if $x$ and $y$ lie in different $C_{i}\,$-categories for at least one $i$
must have $\prod_{i=1}^{N}e(C_{i})$ cells.

We will now see that if $X$ is a product of attributes and each criterion
orders a distinct attribute, then maximal categorization is satisfied and
conversely if maximal categorization is satisfied then we can label
alternatives so that $X$ becomes a product of attributes. \ By joining this
conclusion to result (2), that only binary criteria are efficient, we can
describe efficient decision-making concisely: to be efficient agents must be
able to describe the alternatives in $X$ so that they form a product of
attributes and each criterion must divide a distinct attribute into exactly
two categories.

The simplest way to achieve maximal categorization is for $X$ to be formed
by a product of attributes and for each $C_{i}$ to divide $X$ into
categories based only on attribute $i$. \ The domain $X$ might be a set of
cars, and the attributes might be colors, top speeds, and prices. \ A
`speed' $C_{i}$ would then order cars based on the ranges of top speeds that 
$C_{i}$ deems to be equivalent.

Formally, an \textit{attribute} is a set $X_{i}$ and $N$ attributes define
the domain of alternatives $X=\prod_{i=1}^{N}X_{i}$. \ We will say that a
set of criteria $\mathcal{C}$ \textit{is based on a product of attributes}
if for each $C_{i}$ there is a set $X_{i}$ and a partition of $%
\{X_{i}^{1},...,X_{i}^{e(C_{i})}\}$ of $X_{i}$ such that the categories of $%
C_{i}$ are the sets $X_{i}^{j}\times \left( \prod_{k\neq i}X_{k}\right) $, $%
j=1,...,e(C_{i})$.\ \ So, if $C_{i}$ is an ordering of cars by color then
each $X_{i}^{j}$ would represent a color and $x$ and $y$ would be placed
into distinct $C_{i}$-categories if and only if the $i$th coordinates of $x$
and $y$ indicate different colors: $x_{i}\in X_{i}^{j}$ and $y_{i}\in
X_{i}^{k}$\ where $j\neq k$.\ \ The cells of the discrimination partition $%
\mathcal{P}$\ would then be the $\prod_{i=1}^{N}e(C_{i})$ sets $%
X_{1}^{j_{1}}\times \cdots \times X_{N}^{j_{N}}$ where, for each $i$, $j_{i}$
is an integer between $1$ and $e(C_{i})$. \ Maximal categorization thus
obtains.

This treatment assumes that $X$ is a product space: for each possible
combination of attributes (each possible color-speed-price combination),
there is a corresponding element of $X$. \ But for maximal categorization it
is enough that there is \textit{some} alternative in $X$ for each
combination of attribute ranges specified by the criteria, that is, it is
sufficient for $X$ to be a subset of $\prod_{i=1}^{N}X_{i}$ such that each $%
X_{1}^{j_{1}}\times \cdots \times X_{N}^{j_{N}}$ intersects $X$.

A set of criteria $\mathcal{C}$ that is based on a product of attributes
enjoys a wide-ranging symmetry property. \ Fix some $C_{i}$ in $\mathcal{C}$%
, and consider a set $\mathcal{E}_{-i}$ formed by an arbitrary union of the
categories of the remaining criteria $C_{j}$, $j\neq i$. \ Given the product
structure of $\mathcal{C}$, any such $\mathcal{E}_{-i}$ must intersect each
of the $C_{i}$-categories. \ To continue the car example, the set of cars $%
\mathcal{E}_{-i}$ defined by a certain range of top speeds and prices can be
partitioned into all the possible color subsets, say blue, red, and yellow.
\ If we use $C_{i}$ to order the cells of the color partition of the cars in 
$\mathcal{E}_{-i}$, the ordering will have the same `shape' as -- be order
isomorphic to -- the original color ordering $C_{i}$ of $X$. \ If, for
example, $C_{i}$ on $X$ is a cycle -- blue is better than red which is
better than yellow which is better than blue -- then the $C_{i}$ ordering of
any set of cars defined by a range of speeds and prices will also form a
cycle. \ We conclude that any two sets of cars $Y$ and $Z$ defined by
selections of non-color attributes will be order isomorphic to each other
when each is endowed with the color ordering $C_{i}$ (or rather the
restrictions of $C_{i}$ to $Y$ and $Z$).

This symmetry property may seem to be of limited value since it appears to
apply only to products of attributes. \ But in fact the symmetry property
characterizes any $\mathcal{C}$ that maximally categorizes. \ If for an
arbitrary (possibly nonproduct) $\mathcal{C}$, we apply $C_{i}$ to some $%
\mathcal{E}_{-i}$ and it defines fewer than $e(C_{i})$ $C_{i}$-category
subsets then the discrimination partition $\mathcal{P}$ would have to
contain fewer than $\prod_{i=1}^{N}e(C_{i})$ cells. \ And the only way that $%
\mathcal{E}_{-i}$ and $\mathcal{E}_{-i}^{\prime }$ can each define $e(C_{i})$
$C_{i}$-category subsets is for the $C_{i}$ ordering of these subsets to be
order isomorphic.

Moreover, if an arbitrary (possibly nonproduct) $\mathcal{C}$ enjoys the
symmetry property we can relabel the elements of the domain $X$ so that $%
\mathcal{C}$ is then based on a product of attributes. \ To do this, we
associate each $C_{i}$ with an attribute (e.g., color) and identify each $%
C_{i}$-category with an arbitrary value $X_{i}^{j}$ for that attribute
(e.g., blue): each cell of $\mathcal{P}$\ is thus identified with a vector
of attribute values. \ So, although a product of attributes looks special,
it provides a model for any set of criteria that maximally categorizes.

The following definitions and theorem make these claims precise. \ We use $%
E_{i}^{1},...,E_{i}^{e(C_{i})}$ to denote the categories of criterion $C_{i}$%
.

Given the set of criteria $\{C_{1},...,C_{N}\}$, $\mathcal{E}_{-i}$\ is a 
\textit{union of} $C_{-i}$-\textit{categories} if $\mathcal{E}%
_{-i}=\bigcup_{j}E_{k}^{j}$ for some collection of criterion categories $%
\{E_{k}^{j}\}$ such that $k\neq i$ for each $j$. \ Let $C_{i}^{\mathcal{E}%
_{-i}}$ denote the binary relation defined by $E\,C_{i}^{\mathcal{E}%
_{-i}}E^{\prime }$ if and only if there are $C_{i}$-categories $E_{i}$ and $%
E_{i}^{\prime }$ such that $E=E_{i}\cap \mathcal{E}_{-i}$, $E^{\prime
}=E_{i}^{\prime }\cap \mathcal{E}_{-i}$, and $x\,C_{i}\,y$ for $x\in E$ and $%
y\in E^{\prime }$. \ We then say $\mathcal{C}$ satisfies the \textit{%
order-isomorphism property} if for any $i$ and any two unions of $C_{-i}$%
-categories, $\mathcal{E}_{-i}$ and $\mathcal{E}_{-i}^{\prime }$, the binary
relations $C_{i}^{\mathcal{E}_{-i}}$ and $C_{i}^{\mathcal{E}_{-i}^{\prime }}$
are order-isomorphic.

The set of criteria $\mathcal{C}$ has a \textit{product representation} if
(i) for each $i$, there is a nonempty set $Y_{i}$ and a partition $%
\{Y_{i}^{1},...,$

\noindent $Y_{i}^{e(C_{i})}\}$ of $Y_{i}$, (ii) there is a set of criteria $%
\widehat{\mathcal{C}}=\{\widehat{C}_{1},...,\widehat{C}_{N}\}$ defined on $%
Y=\prod_{i=1}^{N}Y_{i}$ where the categories of each $\widehat{C}_{i}$ are
the sets $Y_{i}^{j}\times \left( \prod_{k\neq i}Y_{k}\right) $, and (iii)
for each $i$, there is a order-preserving bijection $f$ between the
categories of $C_{i}$ and $\widehat{C}_{i}$, that is, $%
E_{i}^{j}C_{i}E_{i}^{j^{\prime }}$ if and only if $f\left( E_{i}^{j}\right) 
\widehat{C}_{i}f\left( E_{i}^{j^{\prime }}\right) $.

We then have the following result.\medskip

\noindent \textit{Theorem}. \ For a set of criteria $\mathcal{C}$, the
following statements are equivalent: (i) $\mathcal{C}$ maximally
categorizes, (ii) $\mathcal{C}$ satisfies the order-isomorphism property,
(iii) $\mathcal{C}$ has a product representation.

\section{Bibliography}

\nocite{*} 
\bibliographystyle{eptcs}
\bibliography{generic}

\end{document}